\documentclass{article}
\PassOptionsToPackage{numbers, compress}{natbib}

\usepackage[preprint]{neurips_2024}
\usepackage{lipsum} 
\usepackage{placeins} 

\usepackage{wrapfig}
\usepackage{amsmath}
\usepackage{multirow} 
\usepackage[utf8]{inputenc} 
\usepackage[T1]{fontenc}    
\usepackage[dvipsnames,table,xcdraw]{xcolor}

\usepackage[pagebackref=false, breaklinks=true, letterpaper=true, colorlinks,
            citecolor=citecolor, linkcolor=linkcolor, bookmarks=false]{hyperref}
\definecolor{citecolor}{HTML}{0071BC}
\definecolor{linkcolor}{HTML}{ED1C24}
\usepackage{url}            
\usepackage{booktabs}       
\usepackage{amsfonts}       
\usepackage{nicefrac}       
\usepackage{microtype}      
\usepackage{graphicx} 

\usepackage{wrapfig}

\title{AID: Adapting Image2Video Diffusion Models  for Instruction-guided Video Prediction}

\author{%
  Zhen Xing\textsuperscript{1}
	\quad
	Qi Dai\textsuperscript{2}
	\quad
	Zejia Weng\textsuperscript{1}
	\quad
	Zuxuan Wu \textsuperscript{1}
	\quad
	Yu-Gang Jiang \textsuperscript{1}
	\\ 
	\textsuperscript{1}Fudan University
	\quad
	\textsuperscript{2}Microsoft Research Asia
    }
\begin{document}

\maketitle

\begin{abstract}

Text-guided video prediction (TVP) involves predicting the motion of future frames from the initial frame according to an instruction, which has wide applications in virtual reality, robotics, and content creation. Previous TVP methods make significant breakthroughs by adapting Stable Diffusion for this task. However, they struggle with frame consistency and temporal stability primarily due to the limited scale of video datasets.
We observe that pretrained Image2Video diffusion models possess good priors for video dynamics but they lack textual control.
Hence, transferring Image2Video models to leverage their video dynamic priors while injecting instruction control to generate controllable videos is both a meaningful and challenging task.
To achieve this, we introduce the Multi-Modal Large Language Model (MLLM) to predict future video states based on initial frames and text instructions. More specifically, we design a dual query transformer (DQFormer) architecture, which integrates the instructions and frames into the conditional embeddings for future frame prediction. Additionally, we develop Long-Short Term Temporal Adapters and Spatial Adapters that can quickly transfer general video diffusion models to specific scenarios with minimal training costs. Experimental results show that our method significantly outperforms state-of-the-art techniques on four datasets: Something Something V2, Epic Kitchen-100, Bridge Data, and UCF-101. Notably, AID achieves 91.2\% and 55.5\% FVD improvements on Bridge and SSv2 respectively, demonstrating its effectiveness in various domains.  More examples can be found at our website \url{https://chenhsing.github.io/AID}.
\end{abstract}

\section{Introduction}
In recent years, diffusion models have made significant advancements in image generation~\cite{stablediffusion, controlnet, dalle}, rapidly becoming one of the hottest topics in computer vision.
The tremendous success of image generation models is based on the collection of billions of high-quality, large-scale, publicly available image-text datasets~\cite{schuhmann2022laion}.  For video generation, Sora~\cite{Sora} and SVD~\cite{svd} achieve remarkable performance fueled by large-scale datasets, which are unfortunately private.

Compared to images, videos encompass dynamic temporal changes and convey richer semantic content~\cite{timesformer,vivit,videoswin, xing2023svformer,wang2024omnivid,openvclip}. Controllable video generation has a broader range of potential applications. Text-guided video prediction (TVP) task is a crucial downstream task in video generation, involving predicting future video frames based on a few initial frames and an instruction text. 
TVP is particularly helpful for generating long videos. Moreover, in robotic and first-person virtual reality (VR) scenarios, predicting future trajectories and states based on the current initial state is highly valuable for robotic arm manipulation and planning in first-person VR devices.

Despite many benefits of the TVP task, it also presents several challenges. These include understanding the initial frames, aligning the initial frames with the instruction text, and generating consistent future frames. Compared to the creative demands of text-to-video models~\cite{hong2022cogvideo, xing2023survey, singer2022make, videoLDM, vdm}, TVP focuses more on the accuracy of the generated video following instructions. Recently, most methods~\cite{gu2023seer, lvdm, videofusion} address this problem by extending text-to-image models~\cite{stablediffusion} to TVP tasks. While these models possess strong creative capabilities, they lack video prior information. Training such models with a small amount of domain-specific data often results in poor video consistency and stability.
For instance, tutorial videos~\cite{somethingdataset}, first-person cooking videos~\cite{epickitchen}, and robotic arm operation videos~\cite{bridgedata} often struggle with generation quality due to the limited scale of video data in these specific fields. However, we observe that general video generation models, which have been pretrained on extensive datasets, have learned robust video dynamics priors. As a result, transferring these well-pretrained video generation models~\cite{svd, Sora, chen2023videocrafter1, singer2022make} to specific domain applications holds significant potential.

\begin{figure*}[t]
\centering
\vspace{-5pt}
\includegraphics[width=0.85\linewidth]{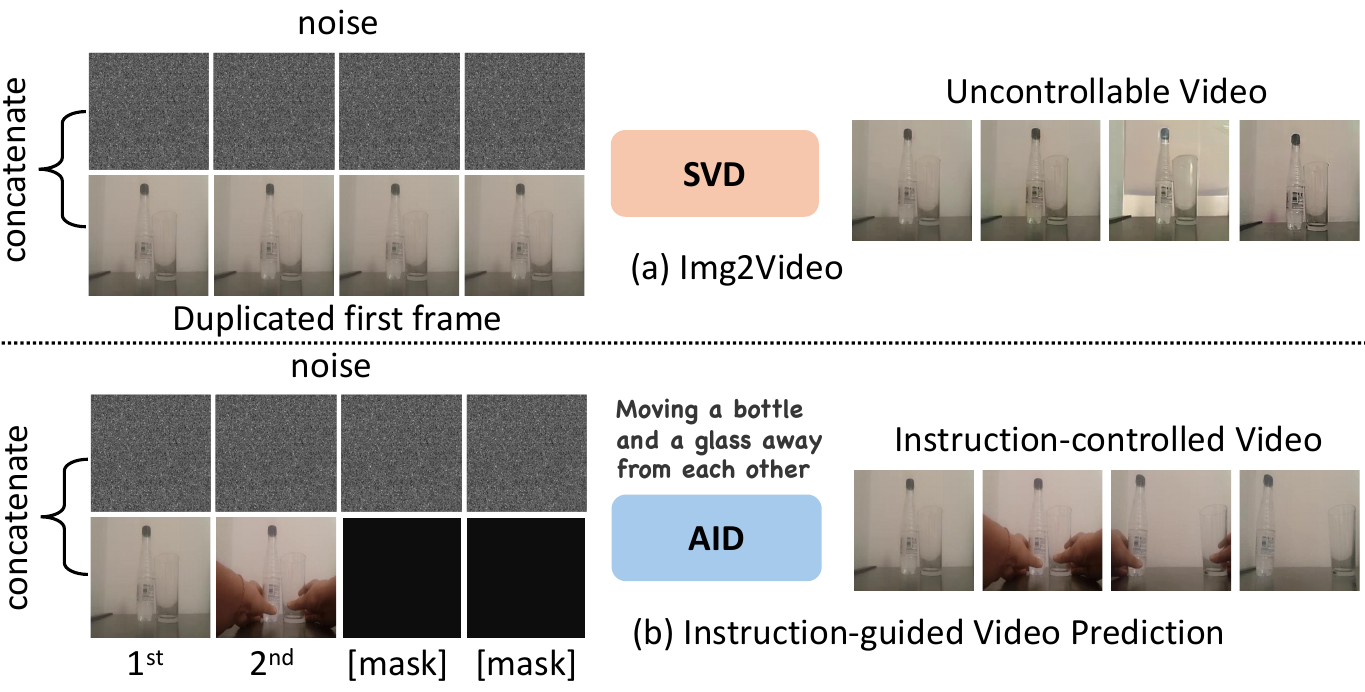}
\vspace{-5pt}
\caption{(a) SVD is an Image2Video generative model that duplicates the initial image and combines this with noises to generate uncontrollable videos through a diffusion model. (b) Our method uses the initial first $K$ frames of the video and a mask of subsequent frames as conditions. Text conditions are also injected into the model to guide the prediction of future frames.}
\vspace{-5pt}
\label{fig:teaser}
\end{figure*}

Among these general video generation models, Stable Video Diffusion (SVD)~\cite{svd} validates the scaling law of pretraining on large-scale video datasets, demonstrating its state-of-the-art performance in video generation.
Thus, we choose it as our base model for image-to-video generation. As illustrated in Figure~\ref{fig:teaser}, although SVD is among the most advanced methods in open-source models, its public version only supports image-to-video generation which is essentially uncontrollable. In practical applications, we prefer to generate video frames based on textual instructions (TVP task). To adapt the Image2Video diffusion model to TVP tasks, we face two main challenges: firstly, how to design the textual condition and inject it into the diffusion model to guide the video generation; secondly, how to adapt the current model to the target dataset at a lower training cost, ensuring that the generated videos more closely to real scenarios.

Predicting future frames based on an initial frame and textual instructions is intuitively challenging, as a single text instruction may not fully capture the temporal dynamics of videos. To address this, we incorporate a Multimodal Large Language Model (MLLM)~\cite{llava} to predict the developmental states of future videos from the initial frame and text instruction. We then design a Dual Query Transformer (DQFormer) architecture with two branches to integrate the existing control information: one branch for learning multi-modal control information from texts and visuals, and the other for decomposing text state conditions into frame-level controls. Finally, the overall multi-condition (MCondition) is injected into the UNet through a cross-attention mechanism. Furthermore, to adapt the model to the target dataset and enhance quality, we design long-term and short-term temporal adapters and spatial adapters, allowing for transferring to the target video prediction with few parameters and computational costs.

We conduct experiments on four mainstream TVP datasets, and the results show that our method significantly surpasses the state-of-the-art~\cite{gu2023seer}. This demonstrates the effectiveness of transferring general generation models~\cite{svd} to specific video prediction tasks. Additionally, the ablation studies further validate the effectiveness of each component in our framework. In conclusion, our main contribution can be summarized as following:
\begin{itemize}
    \item We are the first to transfer an image-guided video generation model to multi-modal guided video prediction, proving the efficacy and vast potential of this approach. 
    \item The proposed DQFormer not only aligns the multi-modal conditions of the initial frames with the instruction but also incorporates the action state conditions predicted by MLLM to guide the video prediction.
    \item  We design the temporal and spatial adapters to achieve training efficiency with few parameters and computational costs.
    \item  Our experimental results demonstrate over 50\% improvement in Fréchet Video Distance (FVD) across multiple datasets compared to previous SoTA.
\end{itemize}

\section{Related Work}

\noindent \textbf{Video Diffusion Models}
Inspired by the substantial success of diffusion models in tasks such as image generation~\cite{stablediffusion, dalle} and image editing~\cite{brooks2023instructpix2pix, controlnet, meng2021sdedit}, research into video diffusion models has shown explosive growth~\cite{xing2023vidiff, zhang2023adadiff,xing2023survey,zhu2024poseanimate}. VDM~\cite{vdm} is the first to apply diffusion to video generation, achieving preliminary results. Subsequent works like Make-A-Video~\cite{singer2022make}, VideoLDM~\cite{videoLDM}, and Imagen Video~\cite{ho2022imagenvideo} focuse on extending T2I (Text-to-Image) models to T2V (Text-to-Video) tasks, achieving significant advancements. Following SimDA~\cite{xing2023simda}, AnimateDiff~\cite{animatediff}, and Tune-a-Video~\cite{tuneavideo}  design efficient training methodologies for video diffusion models, yielding impressive results. Later works~\cite{pixeldance} like VideoCraft~\cite{chen2023videocrafter1, chen2024videocrafter2}, MicroCinema~\cite{wang2024microcinema}, and EMU-video~\cite{emuvideo} focus on designing Image2Video models that enhance the fidelity of video generation. Most recently, SVD~\cite{svd} and Sora~\cite{Sora}, which are data-driven approaches,  demonstrate that scaling up training data can significantly enhance the quality of video generation.

\noindent\textbf{Video Prediction}
Video prediction refers to predicting future frames based on given initial frame conditions. This task has diverse applications, such as creating animations and tutorial videos, and forecasting robotic arm movements. The earliest methods are based on generative adversarial networks (GANs)~\cite{lee2018stochastic, clark2019adversarial}, followed by many studies using auto-regressive Transformer methods~\cite{tats, MAGE} to predict future video frames. Most recently, many works explore video prediction tasks based on diffusion models~\cite{mcvd, gu2023seer, pvdm}. The most similar to our work, Seer~\cite{gu2023seer} employs a text-based video prediction setting and achieves impressive results by designing temporal modules for Stable Diffusion~\cite{stablediffusion}. However, limited by the small scale of video datasets, the generated videos suffer from poor stability and continuity. However, our proposed AID learns an excellent video prior by transferring a large video diffusion model~\cite{svd}, further enhancing the performance of this task and offering a new approach for future work.

\noindent\textbf{LLM in Diffusion Models}
Large language models(LLM) possess robust language understanding and generation capabilities, and many recent works successfully apply LLMs to controllable image generation~\cite{nextgpt, qu2023layoutllm, feng2024layoutgpt, seed}. Concurrently, video generation tasks also explore using LLMs to extend prompts or plan descriptions for multiple video frames. Free-Bloom~\cite{freebloom} and VStar~\cite{li2024vstar} utilize LLMs as a director to break down video prompts into descriptions for each frame, Dysen-VDM~\cite{dysenvdm} employs GPT~\cite{GPT4} to generate dynamic scene graphs that accurately analyze action sequences. Other studies utilize LLMs to predict the layout movement of entities~\cite{lian2023llmground, lv2023gpt4motion, lu2023flowzero}, achieving layout-controllable video generation. Sora~\cite{Sora} also leverages the linguistic capabilities of GPT-4~\cite{GPT4} to rewrite prompts, demonstrating the vast potential of LLMs in applications with video diffusion models.

\section{Method}
\label{sec:method}
In this section, we first introduce the preliminaries of diffusion models in Sec.~\ref{sec:preliminary}. Next, the overview of our method is described in Sec.~\ref{sec:overview}. The architecture for text condition injection is presented in Sec.~\ref{sec:textcondition}. Finally, we discuss the detailed design of three adapters in Sec.~\ref{sec:adapter}.

\subsection{Preliminaries of Diffusion Model}
\label{sec:preliminary}
In this section, we will introduce the continuous-time Diffusion Model~\cite{song2020score, karras2022elucidating} framework briefly.
We begin by identifying the original data distribution denoted by \(p_{\text{data}}(x_0)\) and introduce a modified distribution \(p(x; \sigma)\) which incorporates \emph{i.i.d.} Gaussian noise with variance \(\sigma^2\). For sufficiently large values of \(\sigma_{\max}\), this distribution approximates a Gaussian distribution, \( \mathcal{N}(0, \sigma_{\max}^2)\).
Diffusion models exploit this approximation and commence the denoising process at a high noise level, progressively reducing the noise towards zero. The refinement process is numerically simulated using a continuous-time framework through the solution of the following stochastic differential equation (SDE)~\cite{song2020score}:
\[
dx = -\sigma(t)\theta(t)\nabla_x \log p(x; \sigma(t)) dt,
\]
where the score function \(\nabla_x \log p(x; \sigma)\)~\cite{scorematching} is approximated during training by the model \(s_\theta(x; \sigma)\).
The model \(s_\theta(x; \sigma)\) is designed to approximate the score function, and can be parameterized as:
\[
\nabla_x \log p(x; \sigma) \approx s_\theta(x; \sigma) = \frac{(D_\theta(x; \sigma) - x)}{\sigma^2}.
\]
Here, \(D_\theta\) represents a learnable denoiser, tasked with predicting the clean data \(x_0\). This denoiser is optimized via denoising score matching, described by:
\[
\mathbb{E}_{(x_0,c) \sim p_{\text{data}}(x_0,c),(\sigma,n) \sim p(\sigma,n)} \left[ \lambda_\sigma \|D_\theta(x_0 + n; \sigma, c) - x_0\|^2 \right],
\]
where \(p(\sigma, n) = p(\sigma) \mathcal{N}(n; 0, \sigma^2)\) and \(\lambda_\sigma: \mathbb{R}_+ \rightarrow \mathbb{R}_+\) is a weighting function  and $c$ is an arbitrary conditioning signal.
In our approach, we leverage the EDM-preconditioning framework~\cite{karras2022elucidating} following~\cite{svd} to parameterize the denoiser \(D_\theta\), as follows:
\[
D_\theta(x; \sigma) = c_{\text{skip}}(\sigma)x + c_{\text{out}}(\sigma)F_\theta(c_{\text{in}}(\sigma)x; c_{\text{noise}}(\sigma)),
\]
where \(F_\theta\) represents the network being trained, tasked with predicting the noise-free data.

\subsection{Overview}
\label{sec:overview}
\noindent\textbf{Task Definition}
We first introduce the definition of the task for text-guided Video Prediction (TVP). For a video $\{V\}_{i=1}^{N}$, composed of $N$ frames, we assume that the first $K$ frames and a textual description $t$ are given. The objective is to predict the subsequent $N-K$ frames based on the provided initial $K$ frames and the text description $t$. Suppose we currently have a well-pretrained image-to-video model (SVD~\cite{svd}) capable of automatically generating $N$ frames of video from a given single image. The primary goal is to fully leverage this pretrained generative model to incorporate textual condition information and quickly transfer it to TVP tasks in specific datasets, such as robotic planning~\cite{bridgedata} and first-person perspective cooking~\cite{epickitchen} video datasets.

\begin{figure*}[h]
\centering
\includegraphics[width=1.0\linewidth]{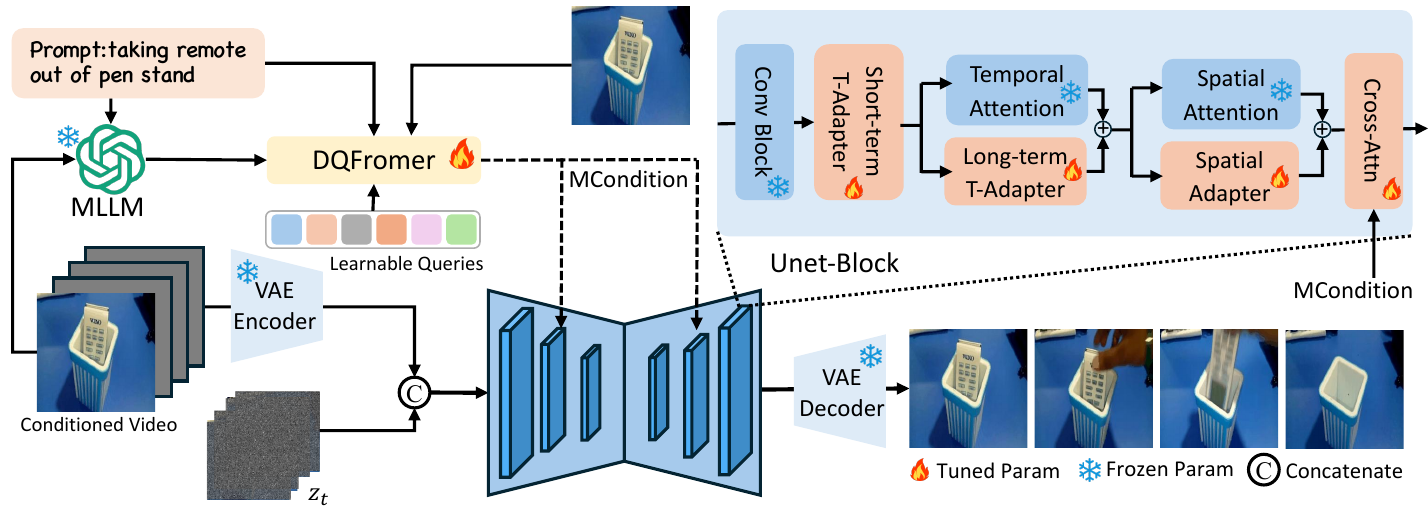}
\caption{(a) The pipeline includes a 3D U-Net for diffusion and a DQFormer for text conditioning. (b) The parameters of the original 3D U-Net are frozen, we only fine-tune the parameters of the newly added adapter during training.}
\label{fig:pipeline}
\end{figure*}

\noindent\textbf{Pipeline }
Our model is built upon Stable Video Diffusion~\cite{svd}, an open-source Image2Video generative model that is pre-trained on large-scale video datasets. As illustrated in Figure~\ref{fig:pipeline}, assuming a video with the first $K$ frames given, we use these as the condition frames. The remaining $N-K$ frames are completed using a mask. After processing through a VAE~\cite{vqvae} encoder to get the conditioned latents, these are concatenated with the noisy latents representation along the channel dimension. Then, controlled by multi-modal condition (MCondition) injected via our specially designed DQFormer, the UNet denoises and predicts the sequence of the $N-K$ latent. Finally, the entire $N$ video frames are reconstructed through the VAE ~\cite{vqvae} Decoder.

\subsection{Text Condition Injection}
\label{sec:textcondition}
In this section, we will discuss how to utilize multimodal large language models (MLLM) to design video prediction prompting, and the architecture of DQFormer for integrating textual and visual conditions into a Multi-Condition (MCondition) to guide video prediction.

\begin{figure*}[h]
\centering
\includegraphics[width=1.0\linewidth]{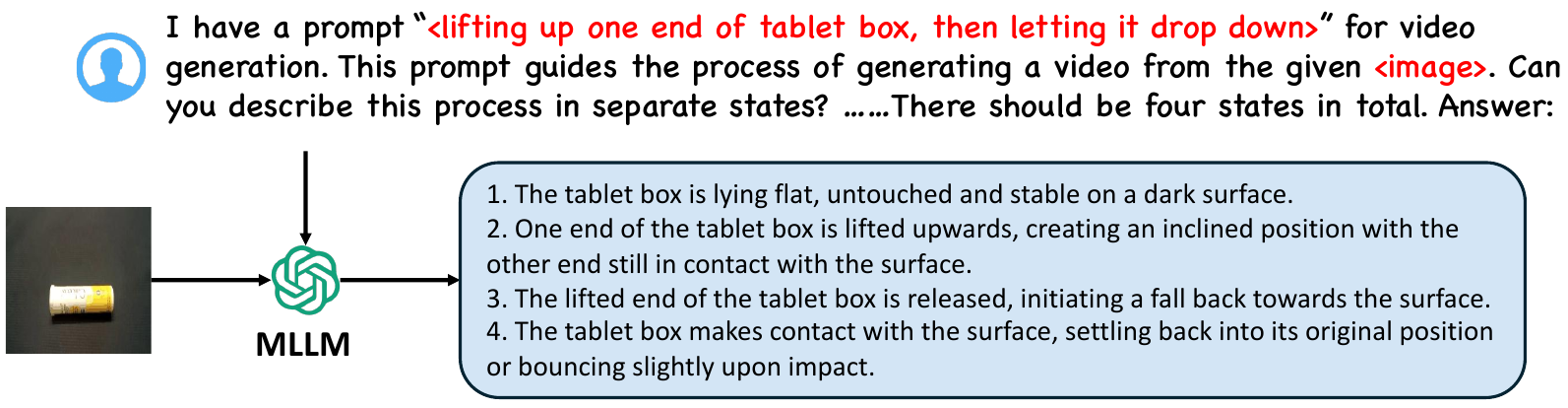}
\caption{We input the initial frame of the video along with the text instruction of the video to be predicted into a multimodal large language model, allowing it to predict multiple state stages of the temporal changes based on the image and text.}
\label{fig:mllm}
\end{figure*}

\noindent\textbf{Video Prediction Prompting}
Text-to-Image models like Stable Diffusion~\cite{stablediffusion} and DALL-E~\cite{dalle} achieve remarkable success, whereas Text-to-Video models are still in development. A significant gap lies in the annotation of text-video datasets, which is more challenging compared to text-image dataset annotation. Moreover, while a single sentence can accurately describe an image, it is insufficient to convey the dynamic changes in a video. Although some methods~\cite{freebloom, li2024vstar} propose using large language models to expand description prompts, existing Text-to-Video models~\cite{animatediff, xing2023simda, videoLDM,singer2022make} typically extract text features from a given sentence and inject the same feature into each frame via cross-attention mechanisms, overlooking the temporal changes in videos. To address this issue, we propose using a multimodal large language model (\emph{e.g.}, LLava~\cite{llava}) to input the initial frame and instruction, allowing it to predict future states of video development. As shown in Figure~\ref{fig:mllm}, for the user prompt "lifting up one end of a tablet box, then letting it drop down," the model can predict four states of the video, including the initial state, lifting one end, releasing one end, and the final state.

\begin{figure*}[h]
\centering
\includegraphics[width=1.0\linewidth]{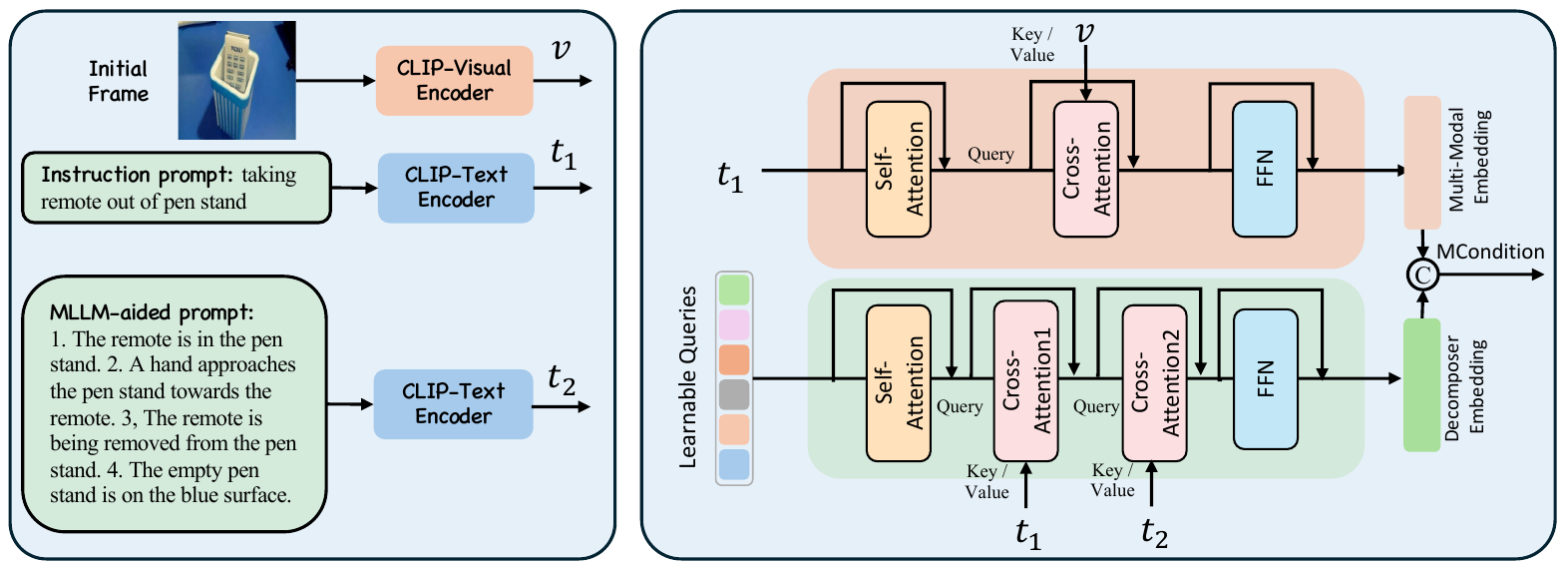}
\caption{For TVP tasks, we design the DQFormer architecture that integrates multiple conditions. The initial frame and two textual prompts are processed by CLIP~\cite{clip} encoder to extract features. The upper branch of DQFormer aligns the visual feature with the textual instruction feature. Meanwhile, the lower branch decomposes the global prompt features into frame-level features. Finally, the MCondition is integrated into each frame of the video through cross-attention.}
\label{fig:dqformer}
\end{figure*}
\noindent\textbf{DQFormer}
With the multi-state texts predicted by the previous MLLM, as well as the initial frame of the video and the instruction prompt, we need to integrate these into a complete multimodal condition 
 (MCondition). To achieve this, we design a Dual Query Transformer (DQFormer) architecture as shown in Figure~\ref{fig:dqformer}, which merges conditional information from various modalities.

We first extract the feature $\boldsymbol{v}$ of the initial frame using the CLIP~\cite{clip} Visual Encoder, and then input the instruction prompt and the MLLM-aided prompt into the CLIP Text Encoder to extract text features $\boldsymbol{t_1}$ and $\boldsymbol{t_2}$, respectively. The right side of Figure~\ref{fig:dqformer} presents our proposed DQFormer, which is inspired by the Q-Former in BLIP-2~\cite{blip2} but features a dual-branch design. 

The upper branch is used for the alignment between textual instruction and the initial frame to obtain $\texttt{multimodal embedding}$. For the global instruction embedding $\boldsymbol{t_1}$, it first passes through multi-head self-attention~\cite{vaswani2017attention}, then computes cross-attention with the visual embedding $\boldsymbol{v}$, and finally obtains the $\texttt{multimodal embedding}$ through a Feed-Forward Network (FFN)~\cite{vaswani2017attention}. The branch can be written formally as following:
\begin{equation}
    \texttt{multimodal embedding} = \texttt{SoftMax}\bigg(\dfrac{(W_1^{Q}\texttt{SelfAttn}(\boldsymbol{t1}))(W_1^{K}\boldsymbol{v})^{T}}{\sqrt{d_1}}\bigg)(W_1^{V}\boldsymbol{v}),
\end{equation}
where $W_1^{Q}$, $W_1^{K}$, and $W_1^{V}$ are learnable parameters and $d_1$ is the scaling coefficient. For clear presentation, we omit the FFN and residual connections.

The lower branch is designed to decompose the prompt features into frame-level condition. Initially, we set up learnable query embeddings $\boldsymbol{Q}\in R^{(N \cdot N_t) \times C}$, where $N$ is the number of frames, $N_t$ is the number of queries which is set to 77 following CLIP~\cite{clip} text encoder.  We first send it to self-attention layer, then compute cross-attention with the instruction embedding $\boldsymbol{t_1}$ to decompose it for each frame. This is followed by computing cross-attention with the MLLM-Aided embedding $\boldsymbol{t_2}$ to decompose the multi-state embedding corresponding to each frame. Finally, the $\texttt{decomposed embedding}$ are obtained through an FFN. The decomposed branch can be represented by the following formula:

\begin{equation}
    \boldsymbol{Q'} = \texttt{SoftMax}\bigg(\dfrac{(W_2^{Q}\texttt{SelfAttn}(\boldsymbol{Q}))(W_2^{K}\boldsymbol{t_1})^{T}}{\sqrt{d_2}}\bigg)(W_2^{V}\boldsymbol{t_1}),
\end{equation}

\begin{equation}
    \texttt{decomposed embedding} = \texttt{SoftMax}\bigg(\dfrac{(W_3^{Q}\boldsymbol{Q'})(W_3^{K}\boldsymbol{t_2})^{T}}{\sqrt{d_3}}\bigg)(W_3^{V}\boldsymbol{t_2}),
\end{equation}

where $W_2^{Q}, W_3^{Q}$, $W_2^{K}, W_3^{K}$, and $W_2^{V}, W_3^{V}$ are learnable parameters and $d_2, d_3$ are the scaling coefficient. Finally, the features are concatenated as  $\texttt{MCondition}$:
\begin{equation}
    \texttt{MCondition} = \texttt{multimodal embedding} \textcircled{c} \texttt{decomposed embedding}.
\end{equation}

After obtaining the $\texttt{MCondition}$, we use it as the key and value, and inject it into the latent representation through the cross-attention mechanism.

\subsection{Adapter Modeling}
\label{sec:adapter}

\begin{wrapfigure}[10]{r}{0.5\textwidth}
\centering
\vspace{-30pt}
\includegraphics[width=1.0\linewidth]{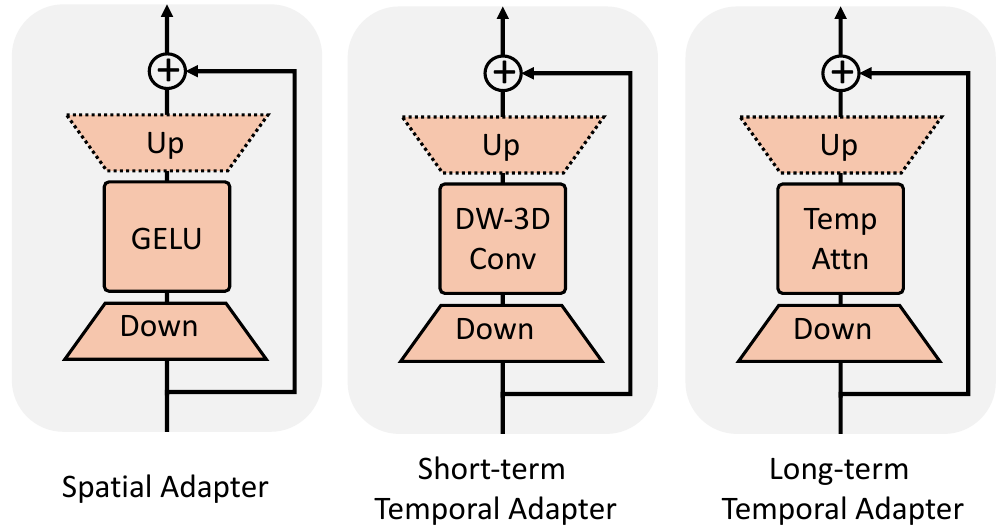}
\caption{The overview of three adapters.}
\label{fig:adapter}
\end{wrapfigure}

In this section, we will detail the three types of adapters we utilize: the spatial adapter, the long-term adapter, and the short-term adapter.

\noindent\textbf{Spatial Adapter}
To adapt the model to the spatial distribution of the target dataset, we design a spatial adapter as shown in Figure~\ref{fig:adapter}, which is added alongside the spatial self-attention. Its structure is simple, consisting of a downsampling linear layer with a GELU~\cite{GELU} activation function and an upsampling linear layer. To ensure that the original model structure is not disrupted, we initialize the upsampling linear layer to zero following ControlNet~\cite{controlnet}. The spatial adapter can be written formally as follows:

\begin{equation}
    \texttt{S\text{-}Adapter}(\mathbf{X}) = \mathbf{X} + \mathbf{W}_{\texttt{up}}(\texttt{GELU}(\mathbf{W}_{\texttt{down}}(\mathbf{X}))),
\end{equation}

\noindent\textbf{Shot-term Temporal Adapter}
Meanwhile, we aim to transfer the dynamic motion distribution of the video to the target dataset. We design two kinds of temporal adapters. The short-term temporal adapter incorporates a Depth-wise 3D Convolution~\cite{depth-wise} between down and up linear layers, which is used for short-term temporal modeling transfer. This can be expressed by the following formula:
\begin{equation}
    \texttt{ST\text{-}Adapter}(\mathbf{X}) = \mathbf{X} + \mathbf{W}_{\texttt{up}}(\texttt{3D-Conv}(\mathbf{W}_{\texttt{down}}(\mathbf{X}))),
\end{equation}

\noindent\textbf{Long-term Temporal Adapter}
The distinction of our designed long-term temporal adapter lies in the incorporation of temporal self-attention between the linear layers. Unlike the convolution-based short-term adapter, which tends to model the temporal relationships between adjacent frames, this adapter is designed to focus on global temporal modeling. Its structure can be described as follows:
\begin{equation}
    \texttt{LT\text{-}Adapter}(\mathbf{X}) = \mathbf{X} + \mathbf{W}_{\texttt{up}}(\texttt{Self-Attn}(\mathbf{W}_{\texttt{down}}(\mathbf{X}))),
\end{equation}
During the training process, we freeze the weights of the original UNet and only update the parameters of the newly added three types of adapters. This approach not only saves GPU memory and training costs but also helps alleviate global overfitting and model collapse.

\section{Experiments}
In this section, we introduce the experimental setup and results of AID in text-guided video prediction tasks. We compared our results with the current state-of-the-art (SoTA) across multiple datasets and validated the effectiveness of the methods mentioned in Sec.~\ref{sec:method} at ablation studies.
\subsection{Datasets}
We perform experiments across three distinct text-video datasets: Something Something-V2 (SSv2)~\cite{somethingdataset}, which features videos of everyday human actions accompanied by verbal instructions; Bridge Data~\cite{bridgedata}, collected from a robotic platform featuring operations of a mechanical arm with textual prompts; and EpicKitchens-100 (Epic100)~\cite{epickitchen}, capturing daily kitchen activities from a first-person perspective with language description. For SSv2, following Seer~\cite{gu2023seer}, we assess the first 2,048 samples from the validation set during evaluations to expedite testing. Bridge Data is divided into an 80\% training segment and a 20\% validation segment for assessments. To simplify, we reduce each video in SSv2 and Epic100 to 12 frames and to 16 frames for Bridge Data during training and evaluation following~\cite{gu2023seer}. Additionally, we further extend our evaluations to include the UCF-101 dataset~\cite{ucf101} in Appendix~\ref{sec:ucf}.

\subsection{Implementation Details}
We initialize the VAE~\cite{vqvae}, Conv block, and Attention block of the U-Net with pre-trained weights from Stable Video Diffusion~\cite{svd}. During the training phase, we fix all layers of both the VAE and the 3D U-Net. We exclusively train our added components: the DQFormer, the three types of adapters, and the Cross-Attention layers. The resolution of the video is resized to $256\times256$ in both train and inference phases. During inference, we employ classifier-free guidance, using two guidance scales designed around frame condition scale $s_v$ and text condition scale $s_t$ to control video generation.  During inference, the noise estimated at time step $t$ is computed as:
\begin{equation}
\resizebox{0.7\linewidth}{!}{
$
\begin{aligned}
     \tilde{e_\theta}(z_t, c_T, c_V) &= e_\theta(z_t, \emptyset, \emptyset) + s_V \cdot (e_\theta(z_t, c_V, \emptyset) - e_\theta(z_t, \emptyset, \emptyset)) \\
     &\quad + s_T \cdot (e_\theta(z_t, c_V, c_T) - e_\theta(z_t, c_V, \emptyset)).
\end{aligned}
$
}
\end{equation}
Where $c_T$ and $c_V$ refer to text condition and frame condition. 

\subsection{Evaluation Setting}
\noindent\textbf{Baselines}: We compare AID with eight baseline methods for video generation, following the Seer~\cite{gu2023seer}. These include: (1) conditional video diffusion methods such as Seer~\cite{gu2023seer}, Tune-A-Video~\cite{tuneavideo}, Masked Conditional Video Diffusion (MCVD)~\cite{mcvd}, Video Probabilistic Diffusion Models (PVDM)~\cite{pvdm}, and VideoFusion~\cite{videofusion}; (2) autoregressive transformer methods like Time-Agnostic VQGAN and Time-Sensitive Transformer (TATS)~\cite{tats}, and Make It Move (MAGE)~\cite{MAGE}; (3) a CNN-based encoder-decoder approach SimVP~\cite{simvp}.

\noindent\textbf{Evaluation Metric}: 
We assess the performance of text-driven video prediction using various baseline methods on several datasets: Something-Something V2 (SSv2)~\cite{somethingdataset} with 2 reference frames, Bridgedata~\cite{bridgedata} with 1 reference frame, and Epic-Kitchens-100~\cite{epickitchen} (Epic100) with 1 reference frame. Additionally, we conduct multiple ablation studies on SSv2 to evaluate the effectiveness of our proposed modules. In our analysis, we use the Fréchet Video Distance (FVD)~\cite{fvd} and Kernel Video Distance (KVD) metrics, calculated using the Kinetics-400 pre-trained I3D model~\cite{kinetic}. We test these metrics on 2,048 samples from SSv2, 5,558 samples from Bridgedata, and 9,342 samples from Epic100 within their respective validation sets following~\cite{gu2023seer}. Specifically, for FVD and KVD, we employ the evaluation methodology from VideoGPT~\cite{yan2021videogpt} and Seer~\cite{gu2023seer}. Moreover, we extend our evaluation to class-conditioned video prediction on the UCF-101~\cite{ucf101} dataset and detail the comparative results in Appendix~\ref{sec:ucf}.

\begin{table}[]
\caption{Text-conditioned video prediction (TVP) results on Something-Something V2 (SSv2), Bridgedata (Bridge), and Epic-Kitchens-100 (Epic100). We report the FVD and KVD metrics of each method in SSv2, Bridge, and Epic100. The result of other methods are duplicated from Seer~\cite{gu2023seer}.}
\centering

\scalebox{0.95}{
\begin{tabular}{@{}ccc|cc|cc|cc@{}}

\toprule
\multirow{2}{*}{\bf Method} & \multirow{2}{*}{\bf Text} & \multirow{2}{*}{\bf Resolution} & \multicolumn{2}{c|}{\bf SSv2} & \multicolumn{2}{c|}{\bf Bridge} & \multicolumn{2}{c}{\bf Epic100} \\ \cmidrule(l){4-9} 
                        &                       &                             & FVD $\downarrow$         & KVD $\downarrow$        & FVD $\downarrow$          & KVD $\downarrow$         & FVD $\downarrow$          & KVD $\downarrow$          \\ \midrule
TATS~\cite{tats}                    & No                    & $128\times128 $                  & 428.1       & 2177       & 1253         & 6213        & 920.0        & 506.5        \\
MCVD~\cite{mcvd}                    & No                    & $256\times256$                     & 1407        & 3.80       & 1427         & 2.50        & 4804         & 5.17         \\
SimVP~\cite{simvp}                   & No                    & $64\times64$                       & 537.2       & 0.61       & 681.6        & 0.73        & 1991         & 1.34         \\
MAGE~\cite{MAGE}                    & Yes                   & $128\times128$                     & 1201.8      & 1.64       & 2605         & 3.19        & 1358         & 1.61         \\
PVDM~\cite{pvdm}                    & No                    & $256\times256$                     & 502.4       & 61.08      & 490.4        & 122.4       & 482.3        & 104.8        \\
VideoFusion~\cite{videofusion}             & Yes                   & $256\times256$                     & 163.2       & 0.20       & 501.2        & 1.45        & 349.9        & 1.79         \\
Tune-A-Video~\cite{tuneavideo}            & Yes                   & $256\times256$                     & 291.4       & 0.91       & 515.7        & 2.01        & 365.0        & 1.98         \\

Seer~\cite{gu2023seer}                    & Yes                   & $256\times256$                  & 112.9       & 0.12       & 246.3        & 0.55        & 271.4        & 1.40         \\ \midrule
AID (Ours)                    & Yes                   & $256\times256$                     & \bf 50.23        & \bf 0.02       & \bf 21.57             & \bf 0.04           & \bf 52.78             & \bf 0.05              \\ \bottomrule
\end{tabular}
}
\label{tab:main}

\end{table}

\begin{figure*}[h]
\centering
\includegraphics[width=1.0\linewidth]{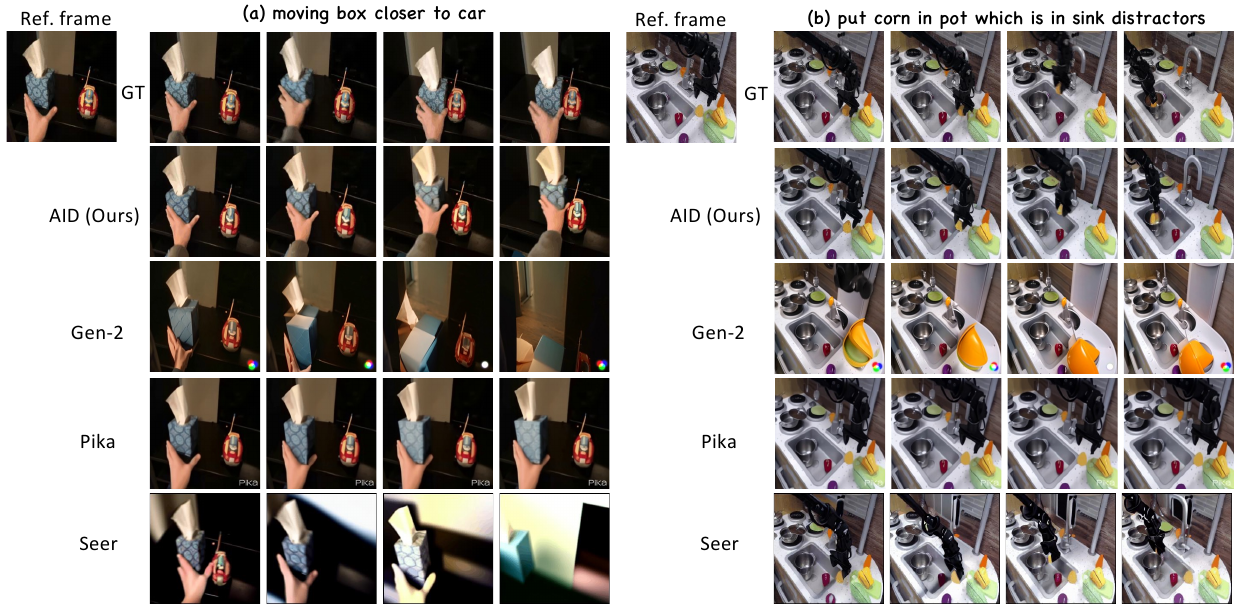}
\caption{Visualization of text-conditioned video prediction:(a) example on Something-Something V2 and (b) example on Bridgedata.}
\label{fig:demo}
\end{figure*}

\subsection{Main Results}
\label{sec:result}
\noindent\textbf{Quantitative Results} 
In Table~\ref{tab:main}, we present the performance of our method, AID, on three datasets in terms of the Fréchet Video Distance (FVD) and Kernel Video Distance (KVD) metrics. We also note whether the baseline methods utilized text conditioning and their resolutions. Specifically, VideoFusion~\cite{videofusion}, Tune-A-Video~\cite{tuneavideo}, and Seer~\cite{gu2023seer} are variants of video diffusion models based on Stable Diffusion~\cite{stablediffusion}, all incorporating text control, aligning with our experimental settings. The table shows that our method significantly surpasses the previous state-of-the-art method Seer~\cite{gu2023seer} across all three datasets. Particularly, in terms of the FVD metric, our method achieves over a 50\% improvement on all datasets, with an outstanding enhancement exceeding 90\% on Bridge Data~\cite{bridgedata}. For instance, on Bridge Data, our FVD has been reduced to just 21.57, indicating that the quality of our generated video is very close to the actual ground truth of real videos.

\noindent\textbf{Qualitative Results}
Figure~\ref{fig:demo} displays a qualitative comparison of our method with the open-source method Seer~\cite{gu2023seer}, as well as two advanced commercial Image2Video models (Gen-2~\cite{gen1} and Pika~\cite{pika}) in the Text-guided Video Prediction task. The case (a) is from a sample in the SSv2~\cite{somethingdataset} dataset, where for the prompt "moving box closer to car," the general models Pika~\cite{pika} and Gen-2~\cite{gen1} fail to understand the action instruction. Seer~\cite{gu2023seer} approach results in unstable frames, with later frames potentially deviating significantly from earlier ones due to cumulative errors. Similarly, the right case shows results from the Bridgedata~\cite{bridgedata} dataset. For specific scenarios like robotic arm video generation, our method demonstrates the strongest text comprehension and video generation stability. For the instruction "put corn in pot which is in sink distractors," our method not only accurately locates the corn but also places it appropriately in the sink distractors. Other methods either misunderstand the instruction or generate unstable videos (\emph{e.g.}, Seer~\cite{gu2023seer} exhibits a ghosting effect with the "corn"). More visualization results are demonstrated in Appendix~\ref{sec:vis_appendix}.

\subsection{Ablation Study}
\label{sec:ablation}

\begin{wraptable}[11]{r}{0.38\textwidth}
\vspace{-25pt}
\centering\small
\setlength{\tabcolsep}{4pt}
\caption{The ablation of different conditions. "DE", "ME", "MC" refer to Decomposed Embedding, Multi-modal Embedding and MCondition respectively.}
\begin{tabular}{@{}ccccc@{}}
\toprule
       & DE & ME & FVD ($\downarrow$)   & KVD ($\downarrow$)  \\ \midrule
w/o MC &    &    & 152.4 & 0.14 \\
w/o DE &    & \checkmark   & 74.98 & 0.03 \\
w/o ME & \checkmark   &    & 70.16 & 0.04 \\
w/o LLava  & \checkmark   & \checkmark    & 64.48 & 0.03 \\
AID (Ours)   &\checkmark    & \checkmark    & \bf 50.23 &\bf 0.02 \\ \bottomrule
\end{tabular}
\label{table:ablation_condition}
\end{wraptable}

In this section, we conduct ablation studies on  SSv2~\cite{somethingdataset}, the largest dataset, to verify the effectiveness of the various components introduced in our method.
More ablations and qualitative comparisons are provided in the Appendix.

\noindent\textbf{Effectiveness of MCondition}
To validate the effectiveness of the MCondition we designed, we conduct ablation studies as shown in Table~\ref{table:ablation_condition}. Our proposed DQFormer consists of two branches: one generates the Multi-Modal Embedding (ME) and the other constructs the Decomposed Embedding (DE). The experimental results indicate that both branches are essential; the absence of either branch (\emph{i.e.}, "w/o DE" or "w/o ME") leads to poorer model performance on both FVD  and KVD metrics. Additionally, if both branches are removed, \emph{i.e.}, training a video prediction model without text guidance ("w/o MC"), the FVD significantly worsens, as it lacks control over the text, and relying solely on the initial frame makes it difficult to predict the state of future videos. Moreover, our ablation studies show that omitting the state prompts predicted by LLava~\cite{llava} ("w/o LLava") also decreases model performance, confirming the benefits of using the MLLM-aided video prompting.

\begin{wraptable}[13]{r}{0.38\textwidth}
\vspace{-20pt}
\centering\small
\setlength{\tabcolsep}{4pt}
\caption{The ablation of different adapters. "SA", "TA", "STA", "LTA" refer to Spatial Adapter, Temporal Adapter, Short-term Temporal Adapter and Long-term Temporal Adapter respectively.}
\scalebox{0.85}{
\begin{tabular}{@{}llllll@{}}
\toprule
        & SA & STA & LTA & FVD($\downarrow$)   & KVD($\downarrow$)  \\ \midrule
w/o Adapter  &    &     &     & 279.42  & 0.71  \\
w/o SA  &    & \checkmark    & \checkmark    & 68.12 & 0.05 \\
w/o TA  & \checkmark   &     &     & 76.32 & 0.03 \\
w/o STA & \checkmark   &     & \checkmark    & 59.62 & 0.03 \\
w/o LTA & \checkmark   & \checkmark    &     & 58.16 & 0.02 \\
AID (Ours)    &\checkmark    &\checkmark     &\checkmark     & \bf 50.23 & \bf 0.02 \\ \bottomrule
\end{tabular}
}
\label{table:ablation_adapter}
\end{wraptable}

\noindent\textbf{Effectiveness of Adapters}
To validate the effectiveness of the three adapters we designed, we conduct ablation experiments as shown in Table~\ref{table:ablation_adapter}. "w/o Adapter" refers to fixing the parameters of U-Net and only training the DQFormer, which results in the most significant decrease in model performance. This proves that by relying solely on the training of a conditional injection module, the model cannot be adapted to transfer to the target dataset. Additionally, we separately valid the removal of the spatial adapter ("w/o SA"), long-term temporal adapter ("w/o LTA"), short-term temporal adapter ("w/o STA") and temporal adapter("w/o TA"). The performance of the model becomes worse in each case, indicating that each of the three adapters we designed plays an effective role in transferring the distribution of videos to specific domains.

\section{Conclusion}
In this paper, we introduced AID for text-guided video prediction tasks. To better predict the future state of videos, we employed a multi-modal large language model for video prediction prompting. Additionally, we designed a dual-branch DQFormer module to integrate control conditions of various modalities. Lastly, we utilized spatial and temporal adapters that enable model transfer with few parameters and training costs. We are the first to explore transferring a well-pretrained video diffusion model to domain-specific video generation tasks, and our experiments demonstrate the feasibility and vast potential of this method, paving the way for future research.

\newpage
{
\bibliographystyle{abbrv}
\bibliography{main}
}

\newpage
\appendix

\section*{Appendix}

\section{Additional Experimental Results}

\subsection{Additional Results on UCF-101}
\label{sec:ucf}
Our method AID, is primarily designed for the text-guided video prediction(TVP) task and is validated on task-level datasets. Most previous text-guided video generation methods~\cite{vdm, singer2022make, magicvideo, lvdm} use UCF-101~\cite{ucf101} as a benchmark for testing. Although class-conditioned video prediction on UCF-101 is not particularly suitable for TVP tasks, we still conduct experiments under two settings to evaluate the video generation performance.

\paragraph{Settings} We fine-tune our model on the UCF-101 dataset, resizing the video resolution to 256x256 with 16 frames per video. We conduct experiments under two settings: predicting videos conditioned on 1 and 5 reference frames following~\cite{hong2022cogvideo, gu2023seer}. We report FVD and FID metrics following the methods of Seer~\cite{gu2023seer} and VDM~\cite{vdm}. During the testing phase, we sample 2,048 samples from the test set following~\cite{gu2023seer}. We perform class-conditioned video prediction on this dataset by writing one sentence for each class as the caption for video generation, following the PYoCo~\cite{pyoco} method. For example, we rewrite "biking" as "A person is riding a bicycle."

\begin{table}[ht]
\centering
\caption{Class-conditioned video prediction performance on UCF-101. We evaluate the AID on the UCF-101 with 16-frames-long videos. Ex.data indicates that the model has been pre-trained or fine-tuned on extra datasets.}
\begin{tabular}{@{}cccccc@{}}
\toprule
\bf Method       &\bf Ex.data              &\bf Cond.  &\bf Resolution &\bf FVD ($\downarrow$)   &\bf FID ($\downarrow$)   \\ \midrule
MoCoGAN~\cite{tulyakov2018mocogan}   & No                   & No & $64 \times 64$    & -   & 26998 \\
MoCoGAN-HD~\cite{mocoganhd}   & No                   & Class. & $256 \times 256$    & 700   & - \\
TGAN-ODE~\cite{tgan_ode}   & No                   & No & $64 \times 64$    & -   & 26512 \\
TGAN-F~\cite{tgan_f}   & No                   & No & $128 \times 128$    & -   & 7817 \\
DIGAN~\cite{DIGAN}        & No                   & No     & -          & 577   & - \\
TGANv2~\cite{tganv2}       & No                   & Class. & $128\times 128$    & 1431  & 3497 \\
VDM~\cite{vdm}          & No                   & No     & $64\times 64$      & -     &295 \\
TATS-base~\cite{tats} & No                   & Class. & $128\times 128$    & 278   & - \\
MCVD~\cite{mcvd}         & No                   & No     & $64\times 64$      & 1143  & -     \\
LVDM~\cite{lvdm}         & No                   & No     & $256\times 256$    & 372   & -    \\
MAGVIT-B~\cite{yu2023magvit}  & No                   & Class. & $128\times 128$    & 159   & - \\
PYoCo~\cite{pyoco}  & No                   & No & $256\times 256$    & 310   & - \\
Dysen-VDM~\cite{pyoco}  & No                   & No & $256\times 256$    & 255   & - \\
VDT~\cite{lu2023vdt}  & No                   & No & $64\times 64$    & 226   & - \\
\midrule
VideoFusion~\cite{videofusion}  & txt-video            & Class. & $128\times 128$    & 173   & - \\
CogVideo~\cite{hong2022cogvideo}     & txt-img \& txt-video & Class. & $160\times 160$    & 626   & - \\
Make-A-Video~\cite{singer2022make} & txt-img \& txt-video & Class. & $256\times 256$    & 81.25 & - \\
MagicVideo~\cite{magicvideo}   & txt-img \& txt-video & Class. & -          & 699   & -     \\ 
AID (1 Ref. frames)         & txt-img \& txt-video & Class. & $256\times 256$    &  102      &  16.5      \\ \midrule
CogVideo~\cite{hong2022cogvideo} (5 Ref. frames) & txt-img \& txt-video              & Class. & $160\times 160$    & 109.23 & - \\
Seer~\cite{gu2023seer} (5 Ref. frames) & txt-img              & Class. & $256\times 256$    & 260.7 & - \\
AID (5 Ref. frames)         & txt-img \& txt-video & Class. & $256\times 256$    & \bf 61.22      & \bf 12.1      \\ \bottomrule
\end{tabular}
\label{tab:ucf}
\end{table}

\begin{figure*}[ht]
\centering
\includegraphics[width=1.0\linewidth]{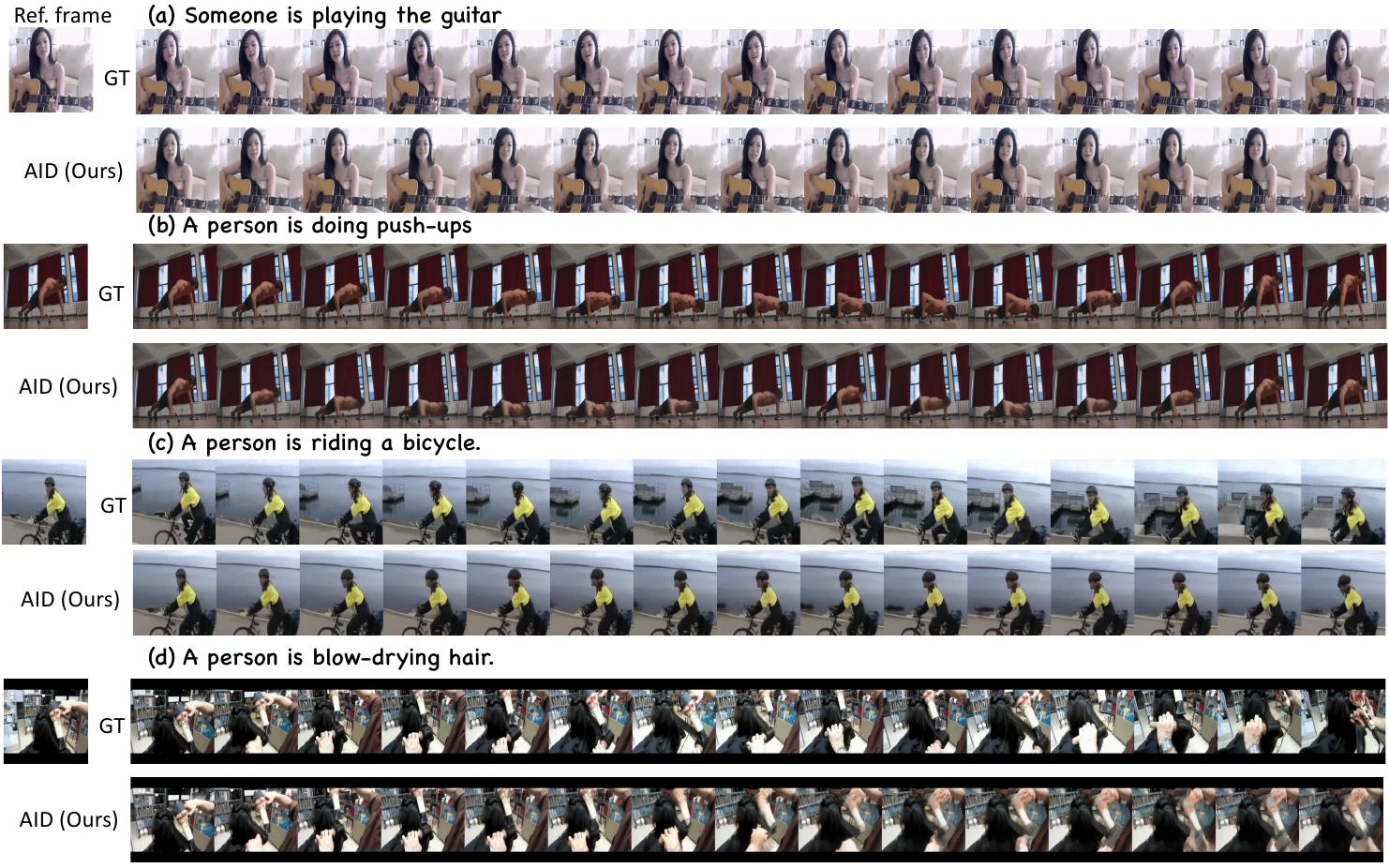}
\caption{Visualization of text-conditioned video prediction on UCF-101 with 1 reference frame.}
\label{fig:ucf1}
\end{figure*}

\begin{figure*}[ht]
\centering
\includegraphics[width=1.0\linewidth]{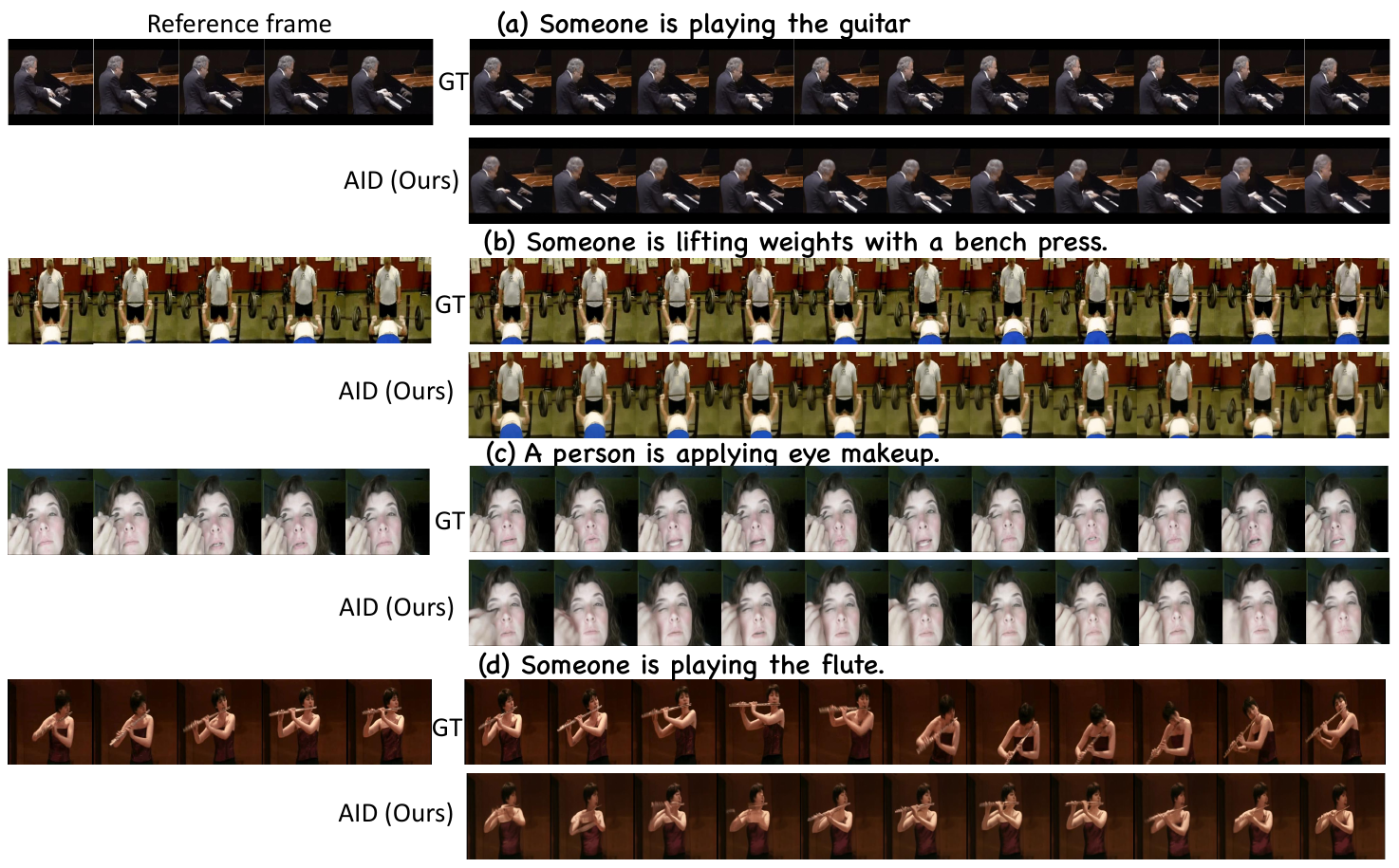}
\caption{Visualization of text-conditioned video prediction on UCF-101 with 5 reference frames.}
\label{fig:ucf5}
\end{figure*}

\paragraph{Results} We present the class-conditioned video prediction results on UCF-101 in Table~\ref{tab:ucf}. When given 1 reference frame, our method significantly outperforms other video generation models in Fréchet inception distance (FID)~\cite{fid} and achieves comparable  Fréchet video distance (FVD)~\cite{fvd} results to the large-scale pre-trained Make-A-Video~\cite{singer2022make} method. For the TVP task with 5 reference frames, our method performs much better than other methods. Additionally, we provide qualitative results in Figure~\ref{fig:ucf1} and Figure~\ref{fig:ucf5}.

\begin{figure*}[ht]
\centering
\includegraphics[width=1.0\linewidth]{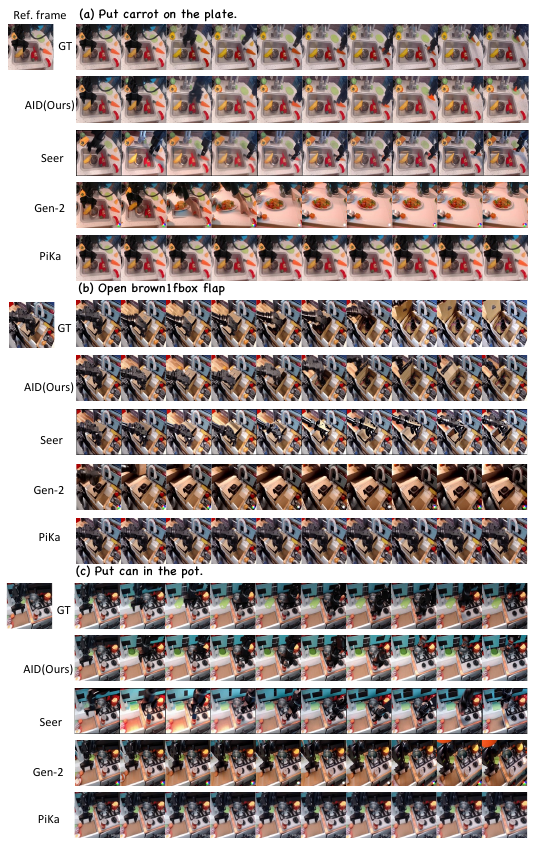}
\caption{Visualization of text-conditioned video prediction on Bridge with 1 reference frame compared to Seer~\cite{gu2023seer}, Gen-2~\cite{gen1}, PiKa~\cite{pika}.}
\label{fig:bridge}
\end{figure*}

\begin{figure*}[ht]
\centering
\includegraphics[width=1.0\linewidth]{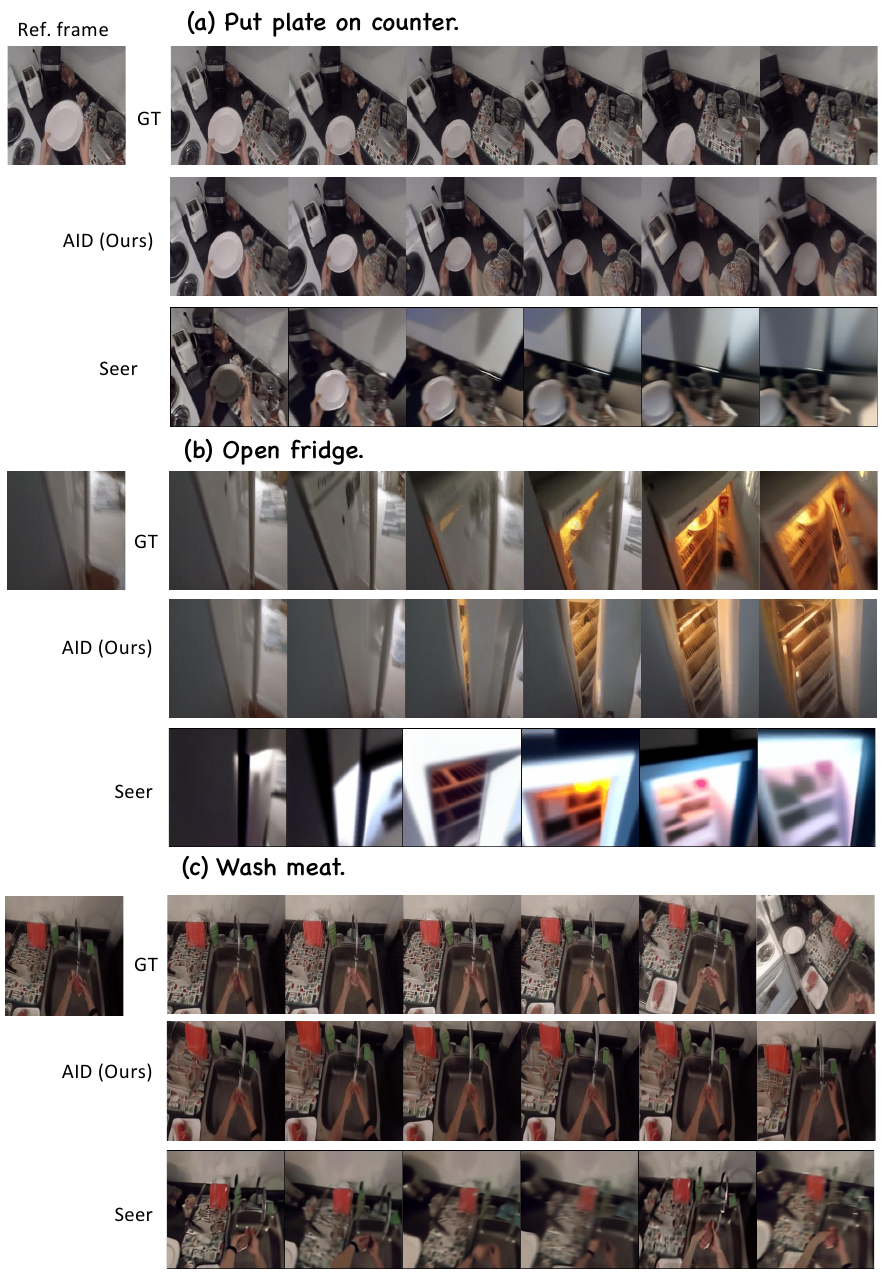}
\caption{Visualization of text-conditioned video prediction on Epic-kitchen with 1 reference frame compared to Seer~\cite{gu2023seer}.}
\label{fig:epic}
\end{figure*}

\begin{figure*}[ht]
\centering
\includegraphics[width=0.9\linewidth]{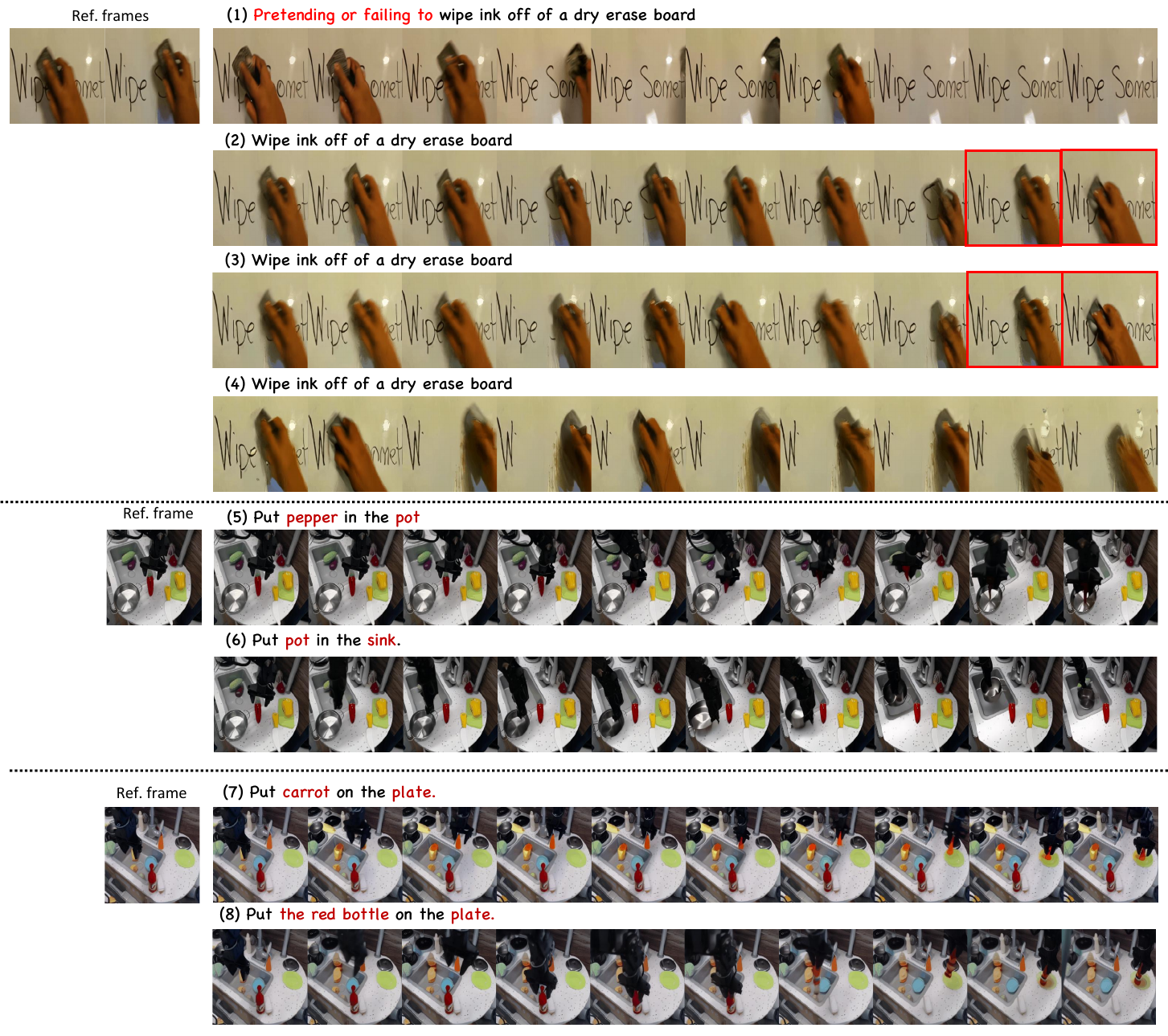}
\caption{Examples of long video prediction and instruction-based video manipulation are provided. }
\label{fig:long}
\end{figure*}

\section{Additional Visualization}
\label{sec:vis_appendix}
In this section, we provide additional visualizations of AID. The results on the Epic-Kitchen dataset are shown in Figure~\ref{fig:epic}, and the results on the Bridge dataset are presented in Figure~\ref{fig:bridge}. We also demonstrate the long video prediction and text-guided video manipulation examples in Figure~\ref{fig:long}. In cases (2)-(4), we use the last two frames of the previous clip as the initial frames for the next clip, allowing iterative extension into longer videos. Other cases demonstrate that given the same reference frame, different instructions can predict different future video frames. Although we provide many sample figures in this paper, we recommend readers visit the link \url{https://chenhsing.github.io/AID} to see the demos in video style.


\end{document}